\documentclass[10pt,twocolumn,letterpaper]{article}

\usepackage{cvpr}              

\definecolor{cvprblue}{rgb}{0.21,0.49,0.74}
\usepackage[pagebackref,breaklinks,colorlinks,allcolors=cvprblue]{hyperref}

\def\paperID{19} 
\def\confName{CVPR}
\def\confYear{2025}
\def\OurDataset{Articulate3D}

\usepackage{subcaption}
\usepackage{float}
\usepackage{graphicx}
\usepackage{booktabs}
\usepackage{xcolor}
\usepackage{comment}
\usepackage{amssymb} 
\usepackage{multirow}
\usepackage{underscore}
\usepackage{amssymb}
\usepackage{pifont}

\title{From Scan to Action:\\Leveraging Realistic Scans for Embodied Scene Understanding}

\author{Anna-Maria Halacheva, Jan-Nico Zaech, Sombit Dey, Luc Van Gool, Danda Pani Paudel\\
\\
INSAIT, Sofia University “St. Kliment Ohridski”
}

\begin{document}
\maketitle
\begin{abstract}
Real-world 3D scene-level scans offer realism and can enable better real-world generalizability for downstream applications. However, challenges such as data volume, diverse annotation formats, and tool compatibility limit their use. This paper demonstrates a methodology to effectively leverage these scans and their annotations. We propose a unified annotation integration using USD, with application-specific USD flavors. We identify challenges in utilizing holistic real-world scan datasets and present mitigation strategies. The efficacy of our approach is demonstrated through two downstream applications: LLM-based scene editing, enabling effective LLM understanding and adaptation of the data (80\% success), and robotic simulation, achieving an 87\% success rate in policy learning.
\end{abstract}
\vspace{-17pt} 
\section{Introduction}
The advancement of technologies that interact with and interpret the real world relies heavily on accurate 3D scene understanding \cite{engelbracht2024spotlightroboticsceneunderstanding,ning2024where2explore}.
Such applications demand rich data, including semantic instance segmentations and articulation annotations. The difficulty in obtaining comprehensive annotations on real-world scans has led to the widespread adoption of synthetic datasets. Yet, while synthetic datasets offer the advantage of large-scale, structured annotations and ease of use, models trained exclusively on them often fail to generalize to the variability and complexity of real-world environments \cite{khanna24hssd,Villasevil-RSS-24}. Conversely, real-world 3D scans capture authentic scene structures and object distributions, but their  fragmented data formats and inherent mesh challenges (e.g., holes, high density), significantly hinder their direct utilization in downstream applications.

This paper explores how to utilize richly annotated real-world 3D scene scans to drive impactful downstream applications, specifically automatic LLM-based scene editing and robotics simulations. We demonstrate how the detailed semantic instance segmentations (object and part level) and articulation annotations from Articulate3D \cite{halacheva2024articulate3d} can be leveraged for these tasks. We present a methodology for unifying these annotations into a USD scene representation, proposing distinct USD flavors optimized for different downstream use cases. The unified format enables easy integration with existing tools and workflows, maximizing the practical utility of real-world scene data.

We provide a detailed methodology for creating two downstream application solutions utilizing holistic real-world scene datasets. First, we introduce an LLM-driven scene editing pipeline. Given a 3D object, its label, and a target scene, our system intelligently integrates the object, leveraging LLMs to determine semantically appropriate placement and scaling. Second, we propose a workflow for creating simulation-ready USD representations of the real-world scene scans, demonstrating their application across diverse control and task scenarios. Notably, we are the first to enable large-scale training for manipulation tasks using real-world scene scans within robotics simulation. We address the inherent challenges of these scans -- including incomplete meshes, object surface gaps, complex geometries, and mesh data overhead -- providing practical solutions for robust and efficient simulation.

Overall, this work makes the following contributions:~ (1) An aggregation appraoch to build interactable USD scenes;
(2) Two flavors of USD;
(3) Demonstration of our approach on two downstream applications.

\section{Method}
We introduce techniques for enabling two downstream applications using 3D open-world scene understanding data. Utilizing the Articulate3D dataset \cite{halacheva2024articulate3d}, which provides comprehensive annotations and complex mesh geometries, we demonstrate how to overcome the challenges associated with leveraging such data for practical applications.

\subsection{Task-Centric Scene Representation}
\label{sec:usd}
Articulate3D \cite{halacheva2024articulate3d} offers a rich set of annotations on ScanNet++ \cite{yeshwanthliu2023scannetppSup} real-world scene scans, enabling holistic scene understanding beyond single segmentation \cite{yeshwanthliu2023scannetppSup,scannet} or isolated articulation prediction tasks \cite{delitzas2024scenefun3d}. It provides object and part-level segmentations for articulated objects, along with labeled fixed, movable, and graspable regions, and articulation parameters, enabling detailed interaction understanding. We demonstrate the effective utilization of these annotations for diverse downstream applications. Articulate3D suggests USD format integration. We found that different downstream applications necessitate distinct USD structure "flavors" for optimal results. We present these application-specific USD flavors, demonstrating their effectiveness.

Universal Scene Description~(USD) scene description format that structures 3D scenes as collections of entities (objects) and relationships, e.g., attachments. Entities can have nested structure, comprising other entities, allowing for object-part hierarchies representation (e.g., a cabinet with doors as sub-entities). 
Additionally, it supports the definition of diverse attributes per entity, including geometric properties, pose, scale, appearance, and user-defined attributes like semantic labels. The format's richness enables varying abstraction levels for different applications. We distinguish between descriptive and geometry-focused USD.

\par\noindent\textbf{Descriptive USD. } USD can describe scene objects, parts, and hierarchies in a script-like format, however, including detailed geometry hinders LLM understanding. For semantic, non-geometry-aware scene understanding, we propose a simplified USD structure containing only object/part labels, bounding boxes derived from segmentations, hierarchies, and articulation data, omitting geometric details.

\par\noindent\textbf{Geometry-Focused USD.} Geometry-aware applications, like physics simulation, require mesh details. However, we observed discrepancies in USD structure interpretation between simulator versions, which we addressed by modifying the hierarchy representation. While object-part hierarchies are useful for script-like representations, they can cause issues like inner collisions and articulation errors in simulations. Therefore, we construct the geometry-focused USD by treating parts as separate objects and signaling hierarchy solely through articulation annotations.

\subsection{LLM-Based Scene Editing}
Even though various datasets with semantic instance segmentation exist, they exhibit limited long-tail object coverage \cite{yeshwanthliu2023scannetppSup,scannet,mao2022multiscan}. Yet, acquiring and annotating new scans requires specialized hardware and significant time investment. An alternative can be provided by scene modifications, e.g., object insertion. To enable intuitive and automated scene modifications, we propose an LLM-driven pipeline for semantically-aware object insertion into real-world indoor scenes. 
The pipeline utilizes the hierarchical and structured representation of USD, enabling LLMs to interpret and edit scenes based on user prompts.

\begin{figure*}[ht]
    \centering
    \includegraphics[width=\linewidth]{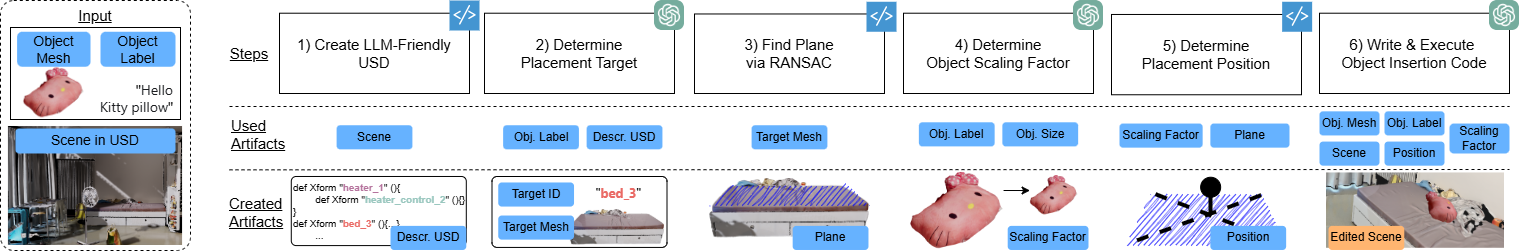}
    \vspace{-15pt}
    \caption{Our pipeline for LLM-based object insertion. It uses two types of USD - scene USD from \cite{halacheva2024articulate3d} and descriptive USD.}
    \label{fig:pipeline}
    \vspace{-11pt}
\end{figure*}

Given a USD scene from \OurDataset{} and its annotations, a 3D object file, and the label of the object to insert, the system generates a modified scene in USD format where the object is placed in a contextually appropriate location. For instance, in a bedroom, a pillow is positioned on a bed, while a water bottle is placed on a desk. Our solution minimizes user involvement in scene editing while demonstrating the capability of LLMs to understand 3D spatial relationships. The editing process is shown in \cref{fig:pipeline}.

\par\noindent\textbf{Object-Insertion Pipeline.} 
Given the inputs, the pipeline first creates an LLM-friendly USD representation of the scene, following the descriptive flavor from \cref{sec:usd}. Specifically, it extracts item IDs, corresponding labels, and connectivity data. Given the descriptive USD, an LLM is prompted to determine the appropriate placement target (e.g., a bed for a pillow) and surface requirements (e.g., pillow - a horizontal plane, poster - a vertical surface). The mesh of the placement target is extracted, and RANSAC is applied to detect a surface for object placement, extracting the relevant surface points. To address scale mismatch between object models and the scenes, the LLM is queried about necessary scaling adjustments based on the object's label and size. The suggested scale, extracted from the response, is applied later in the pipeline. Placement position is determined by calculating the mean of the target surface points and adding a z-offset equivalent to half the object's height. Collision detection is performed, and if collisions occur, a random offset within the target surface's bounds is applied, moving the object along the determined target plane. Placement adjustments and collision checks are iterated until a valid placement is achieved. 
Subsequently, given the determined placement position, scene USD path, object label, object file path and object scaling factor, the LLM generates a USD-Core script for object insertion. This avoids LLM hallucination in the 3D scene and the generated code is executed locally to augment the original USD file.

Recognizing the potential security risks of direct LLM-generated code execution, we implement minimal security checks. We define an allowlist of USD-Core operations and discard any generated code utilizing non-allowlisted operations. Furthermore, we verify that all allowlisted operations originate from the USD-Core library via import statement checks, reducing the risk of introducing vulnerabilities. 

\subsection{Robotics Simulation}
\label{sec:methodsimulation}

Simulation environments predominantly rely on synthetic 3D assets \cite{ning2024where2explore}, ranging from simplified tabletop setups and isolated object representations for basic tasks, to complete, full-scene simulations for enhanced realism and complex manipulations \cite{robocasa2024,Villasevil-RSS-24,procthor}. Although real-world scans are being explored, object interaction tasks remain restricted to single object scans \cite{chu2023command,Villasevil-RSS-24,qiu2025articulateanymeshopenvocabulary3d}. This reliance on synthetic and on single-object real-world assets introduces significant limitations. 
Synthetic assets, by design, feature a restricted set of object classes, oversimplified geometries, and limited articulation mechanisms \cite{kawana2024detection}. Consequently, models trained within these environments often struggle to generalize to the complexities of real-world scenarios \cite{Villasevil-RSS-24,Weng2024CVPR}.
Conversely, while real-world single object scans offer improved geometric fidelity, they present their own set of challenges. The isolated nature of these scans results in observation spaces that differ substantially from real-world application environments, which are characterized by numerous distractor objects and complex scene layouts \cite{ArtObjSim}. 

As an alternative, we present the first adoption of real-world scene-centric scans for manipulation simulation. 

\par\noindent\textbf{Simulation Pipeline.} Our methodology utilizes the Articulate3D \cite{halacheva2024articulate3d} annotations for both movable parts and grasping regions, as well as the articulation parameter. To automate asset preparation for manipulation simulation, users specify a target object ID (e.g., "drawer\_7") and the related articulation and corresponding grasping regions are automatically extracted from the annotations. The grasping position is then determined by computing the mean of the annotated grasping region. This information, together with a simulation-ready USD are sufficient for policy learning.

The construction of a simulation-ready USD scene, a known challenge \cite{Villasevil-RSS-24}, was implemented following the asset structure from \cite{Villasevil-RSS-24}. This process revealed several key challenges inherent in simulating real-world scans:

\textbf{1. Collision Detection:} The high-fidelity meshes from ScanNet++ present complex geometries, diverse object designs, and mesh inconsistencies, leading to incompatibility with convex hull collision detection. We address this issue by utilizing the convex decomposition scheme.

\textbf{2. Object Stabilization:} To prevent manipulated objects from falling, they are fixed to the ground, mirroring the approach in \cite{Villasevil-RSS-24}. This is required since incomplete object meshes resulting from scan limitations, such as missing bottom or back surfaces, lead to unbalanced centers of mass. Consequently, external forces applied during policy learning can destabilize these objects.

\textbf{3. Mesh Decimation for Accelerated Training:} The high-detail meshes require mesh quadratic decimation to enhance training efficiency in scene-level simulations. By retaining 10\% of the original faces for walls, floors, and ceilings, and 30\% for other static objects, we enable parallel environment learning. The manipulated object remains at full resolution to preserve geometric fidelity and interaction accuracy. The decimation strategy is solely applied for policy learning to enable faster learning.

\textbf{4. Mesh Integration:} Due to the scanning process capturing entire scenes rather than individual objects, surfaces beneath placed objects are not recorded. To prevent object displacement due to gravity, we merge meshes of static objects into a single mesh. For pick-and-place tasks, objects of interest need to be translated onto a valid surface.


\section{Experiments}
\subsection{LLM-Based Scene Editing}
We implement the LLM-guided object-insertion pipeline using system instruction prompting and few-shot prompting to achieve improved task performance of the LLM without a requirement for costly finetuning.
We test two LLMs - GPT-4o mini \cite{gpt40mini} and GPT-4o \cite{gpt4o}, observing successful object insertion across both models.
\begin{figure}
  \centering
  \begin{subfigure}[b]{0.115\textwidth}
    \centering
    \includegraphics[width=\linewidth]{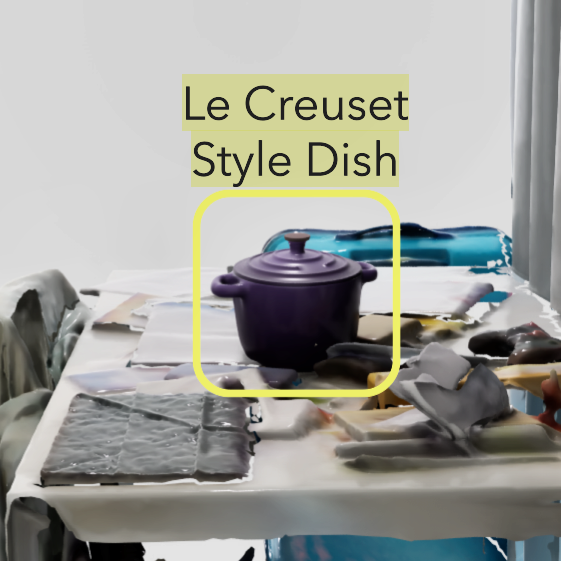}
    \label{fig:subfig1}
  \end{subfigure}
  \hfill
  \begin{subfigure}[b]{0.115\textwidth}
    \centering
    \includegraphics[width=\linewidth]{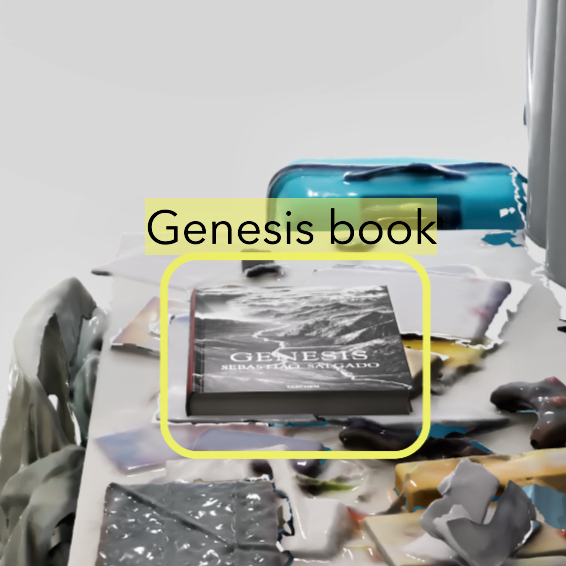}
    \label{fig:subfig2}
  \end{subfigure}
  \hfill
  \begin{subfigure}[b]{0.115\textwidth}
    \centering
    \includegraphics[width=\linewidth]{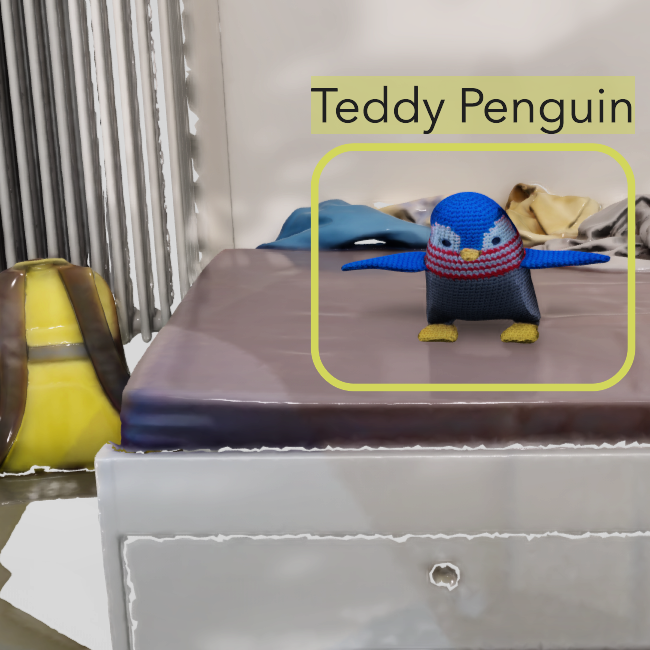}
     \label{fig:subfig3}
  \end{subfigure}
  \hfill
  \begin{subfigure}[b]{0.115\textwidth}
    \centering
    \includegraphics[width=\linewidth]{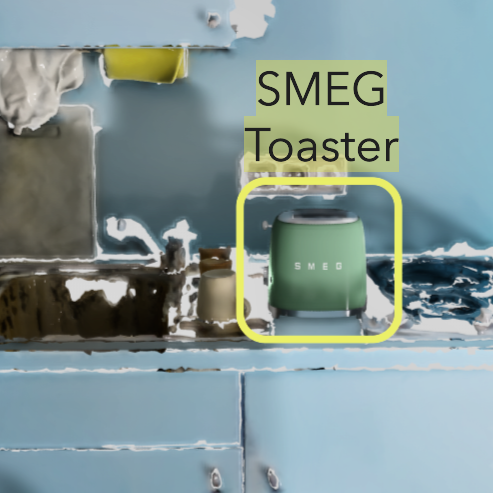}
     \label{fig:subfig4}
  \end{subfigure}
  \vspace{-30pt}
  \caption{Qualitative results for object insertion using LLM-based pipeline. The inserted objects with their labels are marked.}
  \label{fig:insertresults}
  \vspace{-16pt}
\end{figure}

We quantitatively evaluated our approach through a user study with insertions of  100 relevant objects. The objects, sourced from Objaverse XL \cite{deitke2023objaverse}, cover 20 instances from each of the following categories: bathroom objects (personal care and hygiene products), kitchen objects (incl. food), bedroom objects (comfort, decor), stationery, and living room decorations. For evaluation, participants were presented with visualizations of the modified scenes, similar to \cref{fig:insertresults}. They were tasked with classifying each insertion as either a "success" or "failure" with "success" cases meeting the following criteria: semantic appropriateness of placement, accurate object scaling, and absence of mesh collisions. Perfect success (20/20) was achieved in all categories except the bathroom category, which had zero successes, leading to an overall success rate of 80\%. The failure cases can be traced back to unrealistic object placement on the challenging sink geometry and incorrect scaling caused by rare or unclear labels (e.g., brand and product names).

\subsection{Robotics Simulation}
\label{sec:experimentssim}
We conduct experiments employing both planner-based policies and reinforcement learning via Proximal Policy Optimization (PPO) \cite{ppo2017Schulman}, validating our approach with a range of robotic learning paradigms and task requirements. 
Notably, executing planner-based policies enables efficient and scalable data collection, including trajectories, observations, and realistic RGB images, especially beneficial when using full scenes instead of isolated objects.

\par\noindent\textbf{Tasks.}
We consider robotic manipulation tasks to interact with both articulated and inserted objects. For the articulated object interaction task, we train a policy to open a cabinet drawer within an office scene. We demonstrate both the end-to-end policy, trained using PPO, and a planner-based policy.
The simulation setup is fully automated, as described in \cref{sec:methodsimulation}, requiring only the movable part ID.
We further demonstrate a pick-and-place task in a bedroom scene, where the robot grasped and relocated a teddy bear, using planner-based policies. In this scenario, the user can configure the object (e.g., teddy bear) and its grasping position. The object's placement location is determined by an LLM, following our object insertion pipeline.

\begin{figure}
  \centering
  \begin{subfigure}[b]{0.23\textwidth}
    \centering
    \includegraphics[width=\linewidth]{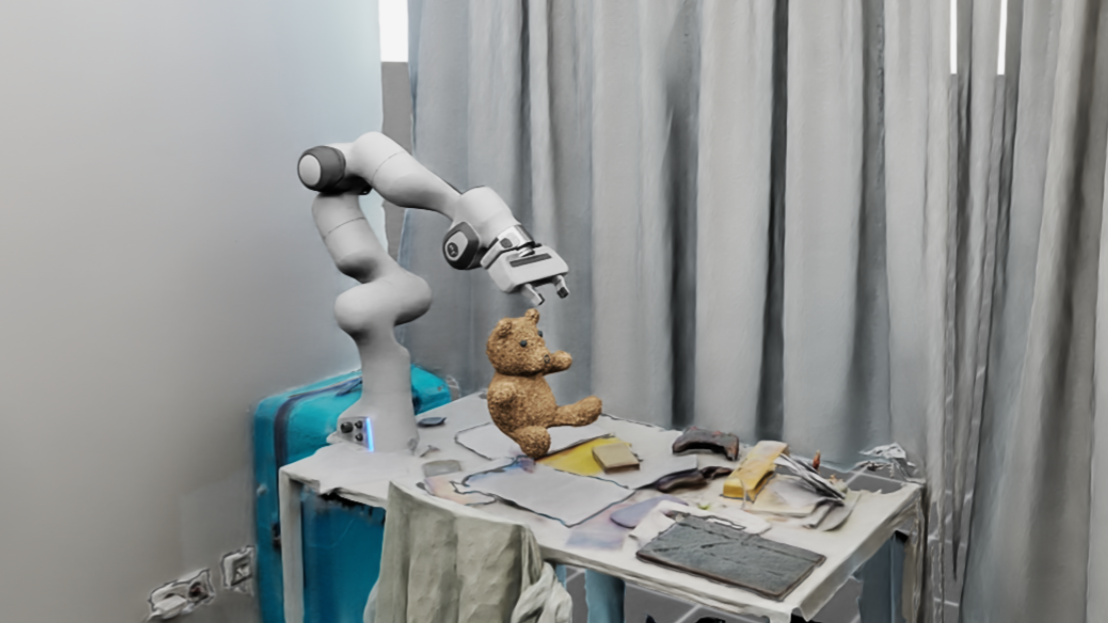}
  \end{subfigure}
  \hfill
  \begin{subfigure}[b]{0.23\textwidth}
    \centering
    \includegraphics[width=\linewidth]{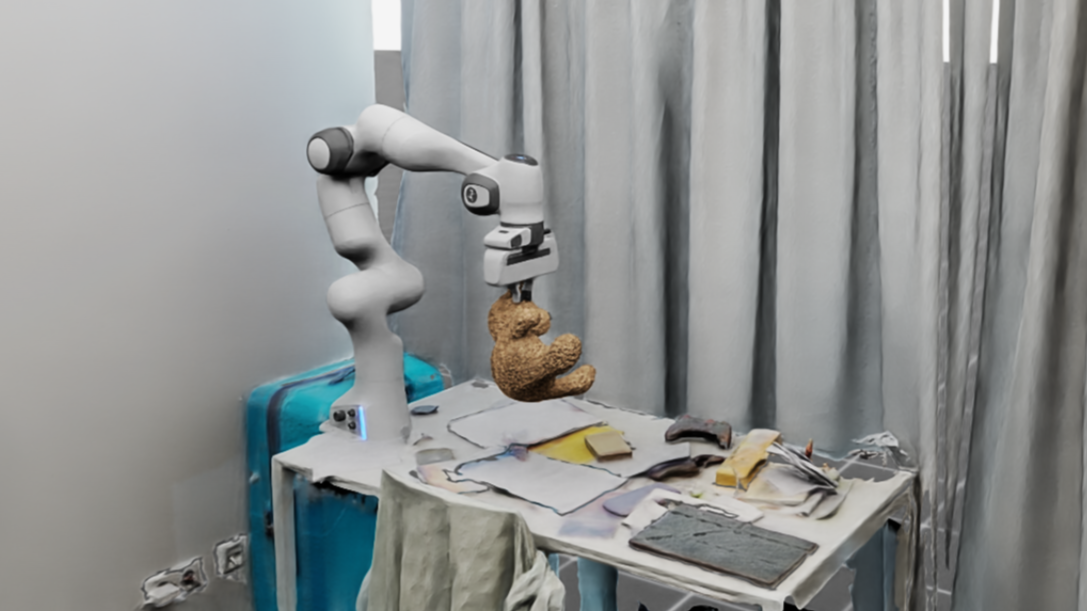}
  \end{subfigure}
  \vspace{-5pt}
  \caption{Pick and place task execution via \textbf{planner-based} policy.}
  \label{fig:pickplace}
  \vspace{-15pt}
\end{figure}

\par\noindent\textbf{Experimental Setup.}
For our simulation experiments, we utilize IsaacSim with IsaacLab \cite{mittal2023orbit}, for their native USD support. We conduct all simulations on a RTX 3090 GPU, employing the Franka Emika robot. The robot is positioned at 0.55m from the target object, within Franka's 0.85m reach, allowing well defined grasping poses.
The policy learning experiments use 1024 environments and are trained for 30,000 iterations with 40 steps per environment, employing a learning rate of 5.0e-4 and an adaptive learning rate schedule.
Drawer opening and pick-and-place tasks used an office and bedroom scene from Articulate3D, respectively, for semantic relevance.

\par
\begin{figure}
  \centering
  \begin{subfigure}[b]{0.23\textwidth}
    \centering
    \includegraphics[width=\linewidth]{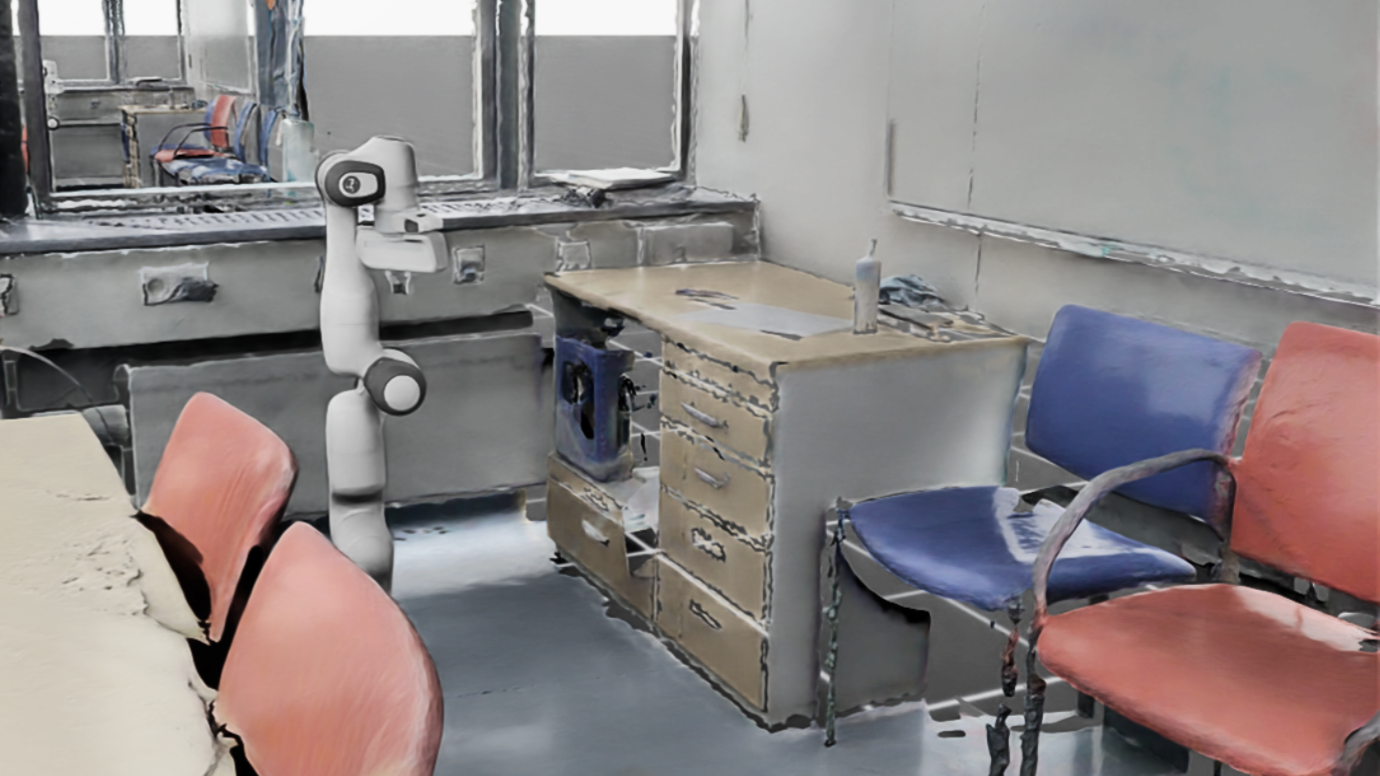}
  \end{subfigure}
  \hfill
  \begin{subfigure}[b]{0.23\textwidth}
    \centering
    \includegraphics[width=\linewidth]{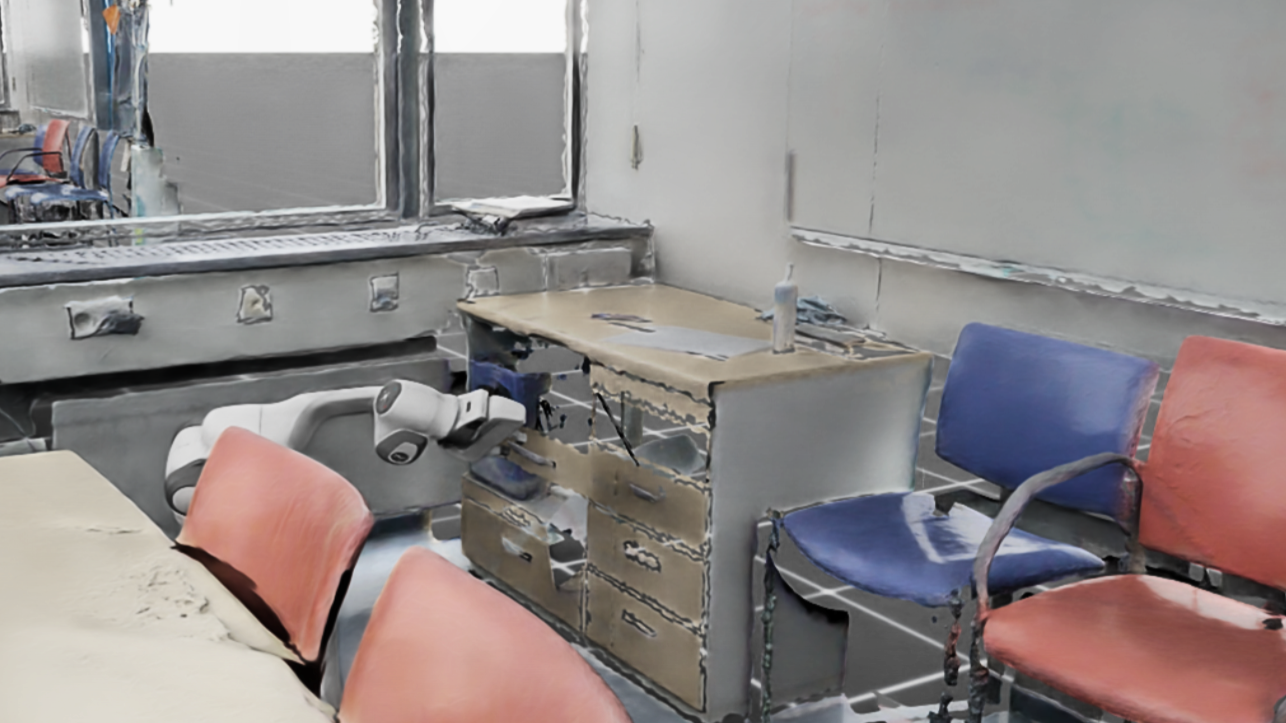}
  \end{subfigure}
  \vspace{-5pt}
  \caption{Drawer opening using a \textbf{PPO-learned} policy.}
  \label{fig:policy}
  \vspace{-15pt}
\end{figure}

\noindent\textbf{Results.}
Learning a drawer-opening policy via PPO, we achieve an 87\% success rate
(success defined as drawer opening $\geq$ 0.2m), with qualitative results shown in \cref{fig:policy}.
The planner-based policies achieved 100\% success rate on all objects interacted with, both rigid and deformable. Pick-and-place qualitative results are offered in \cref{fig:pickplace}.

\par\noindent\textbf{Reward Engineering.} We observed a performance gap during PPO drawer opening experiments between simulation-ready synthetic assets and real-world objects (Articulate3D \cite{halacheva2024articulate3d}), particularly regarding handle designs. To investigate this, we conducted an ablation study on success rate and handle utilization, varying reward functions (\cref{tab:ablation}). Using IsaacLab's PPO drawer-opening reward function and training on their readily available synthetic asset, we achieved a 98\% success rate across 16,384 environments for 1,000 iterations, with consistent handle utilization. Applying this reward function to the real-world cabinet resulted in an 87\% success rate, with Franka frequently grasping the drawer from the side, bypassing the handle. 

We increased the rewards for handle utilization and decreased the opening reward. This encouraged handle use but also led to a decrease to 74\% success rate, as Franka occasionally lost its grasp during the pulling motion. This issue, absent in synthetic simulations, highlights the geometric complexity of real-world data. Manual adjustments to the grasping point did not resolve this, emphasizing the geometric differences.

\begin{table}[h]
\scriptsize
 \vspace{-5pt}
    \centering
    \begin{tabular}{@{\hskip 0.5mm}l@{\hskip 2.8mm}c@{\hskip 2.8mm}c@{\hskip 2.8mm}c@{\hskip 2.8mm}c@{\hskip 2.8mm}c@{\hskip 2.8mm}c@{\hskip 0.5mm}}
        Asset & Success Rate & Handle & Distance & Rotation & Opening & Finger \\
        \midrule
        Synthetic  & 98\%  & \textcolor{green}{\checkmark}& 1.5  & 1.5  & \textbf{10.0}  & 2.0  \\
        Real \cite{halacheva2024articulate3d} & 88\% & \textcolor{red}{$\times$} & 1.5  & 1.5  & \textbf{10.0}  & 2.0  \\
        Real  \cite{halacheva2024articulate3d} & 74\% & \textcolor{green}{\checkmark} & \textbf{50.5} & \textbf{50.5} & 10.0  & \textbf{50.0 } 
    \end{tabular}
    \vspace{-8pt}
    \caption{Ablation study on drawer opening policy training. Reported: asset, success rates, whether the handle was used, reward components contributions (include distance to grasping point, gripper rotation alignment, drawer opening distance, and gripper finger distance). Highest reward component marked in bold.}
    \label{tab:ablation}
     \vspace{-5pt}
\end{table}

\par\noindent\textbf{Observations and Limitations.}
Our experiments demonstrated that \OurDataset~scenes are suitable for robotic simulation without immediate collision or placement issues. However, we acknowledge that certain scenes with specific geometric configurations (e.g., excessively short tables) may present challenges.

\section{Conclusion}
We demonstrated two downstream applications of richly annotated real-world scene scans, emphasizing the importance of holistic annotations. We unified fragmented annotations using the USD format, introducing two specialized USD flavors tailored for semantic descriptiveness and geometric detail. We presented an implementation of an LLM-based scene editing pipeline and robotic simulation for real-world scans, detailing encountered challenges and their solutions. 
We validated our scene editing methodology through a user study, achieving an 80\% success rate, with failures limited to  bathroom objects. In robotic simulation, we achieved an 87\% success for manipulation policy training and 100\% for planner-based policies.

\section*{Acknowledgments}
This research was partially funded by the Ministry of Education and Science of Bulgaria (support for INSAIT, part of the Bulgarian National Roadmap for Research Infrastructure).
{
    \small
    \bibliographystyle{ieeenat_fullname}
    \bibliography{main}
}


\end{document}